\documentclass{article} 
\usepackage{iclr2018_conference,times}
\usepackage{hyperref}
\usepackage{url}

\usepackage[utf8]{inputenc}
\usepackage{amsmath,amsthm,amsfonts,amssymb,euscript,stmaryrd,graphicx,enumitem,yhmath,mathrsfs}

\makeatletter
\newsavebox\myboxA
\newsavebox\myboxB
\newlength\mylenA

\newcommand*\xoverline[2][0.75]{%
    \sbox{\myboxA}{$\m@th#2$}%
    \setbox\myboxB\null
    \ht\myboxB=\ht\myboxA%
    \dp\myboxB=\dp\myboxA%
    \wd\myboxB=#1\wd\myboxA
    \sbox\myboxB{$\m@th\overline{\copy\myboxB}$}
    \setlength\mylenA{\the\wd\myboxA}
    \addtolength\mylenA{-\the\wd\myboxB}%
    \ifdim\wd\myboxB<\wd\myboxA%
       \rlap{\hskip 0.5\mylenA\usebox\myboxB}{\usebox\myboxA}%
    \else
        \hskip -0.5\mylenA\rlap{\usebox\myboxA}{\hskip 0.5\mylenA\usebox\myboxB}%
    \fi}
\makeatother

\usepackage{algorithm,mathtools}
\usepackage{bm}

\usepackage{listings}
\usepackage[english]{babel}
\usepackage{calc}
\usepackage{url}
\usepackage{cleveref}
\usepackage{footmisc}
\usepackage[expansion=false]{microtype}
\usepackage{todonotes}
\usepackage{array}
\usepackage{natbib}
\usepackage{tikz}

\usepackage{todonotes}

\usepackage[noend]{algpseudocode}

\hypersetup{  
	colorlinks=true,
	linkcolor=blue,
	citecolor=blue,
	urlcolor=blue,
	pdfsubject={Complexity Gradient Descent}
}

\newtheorem{theorem}{Theorem}

\theoremstyle{definition}

\title{Combining learning rate decay and weight \\ decay with complexity gradient descent \\ - Part I}

\author{Pierre H. Richemond\\
Data Science Institute\\
Imperial College\\
London, SW7 2AZ\\
\texttt{phr17@imperial.ac.uk}\\
\And
Pr. Yike Guo, Director\\
Data Science Institute\\
Imperial College\\
London, SW7 2AZ\\
\texttt{y.guo@imperial.ac.uk} \\
}

\iclrfinalcopy 

\begin{document}
\maketitle

\begin{abstract}
The role of $L^2$ regularization, in the specific case of deep neural networks rather than more traditional machine learning models, is still not fully elucidated. We hypothesize that this complex interplay is due to the combination of overparameterization and high dimensional phenomena that take place during training and make it unamenable to standard convex optimization methods. Using insights from statistical physics and random fields theory, we introduce a parameter factoring in both the level of the loss function and its remaining nonconvexity: the \emph{complexity}. We proceed to show that it is desirable to proceed with \emph{complexity gradient descent}. We then show how to use this intuition to derive novel and efficient annealing schemes for the strength of $L^2$ regularization when performing standard stochastic gradient descent in deep neural networks.
\end{abstract}

\section{Introduction}
While deep learning optimization has achieved impressive robustness and scale in empirical applications, its theoretical properties are still to date much less understood. And to wit, in spite of abundant research for new stochastic optimizer methods, few of them have consistently outperformed standard methods such as ADAM and variants (\cite{Adam, adamw}) or momentum SGD (\cite{pmlr-v28-sutskever13}). Theoretical difficulties abound when doing analysis on the convergence of gradient descent algorithms, linked both to the non-linearity (and non-convexity) of neural networks losses considered, and the high-dimensional, overparameterized regime in consideration. Very recent work has been able to prove the theoretical convergence of gradient descent for neural networks (\cite{duSGD, zouSGD}), but the path dependency of these methods remains elusive, which is an issue when it comes to leveraging those proofs' insights in order to devise new optimizers. Several angles have been taken in order to remediate this state of affairs - some of the most promising being analyzing the trajectory of gradient descent updates in weight space, as well as gaining insights on the loss landscape of neural networks itself (\cite{choromanska, pratikloss, littwin, essentiallynobarriers}).

In this work, we take the latter view, and revisit methods from complex systems towards understanding the landscape of losses. It is our view that some of the more surprising aspects of deep learning optimization, like the large number of relatively benign local minima or the existence of threshold energies, are better understood and quantified through the prism of statistical physics. Specifically, we use results on Gaussian random fields from \cite{brayanddean, auffingercomplexity}. Our contribution is to recast those in the deep learning optimization framework, and to use them to posit that the \emph{complexity} (as defined in what follows) of the neural network loss function is a relevant surrogate optimization objective which includes valuable second-order loss information besides loss level. We highlight the connection with $L^2$ regularization that arises then. In turn, this enables us to compute its gradient and to derive several alternative optimization algorithms to stochastic gradient descent.

\section{Background and related work}
\textbf{Statistical physics and Hamiltonians.} We begin this section with the analogy with the \emph{dean's problem}. Consider having to organize $N$ students in dorms, where students are either friends or enemies. The dean's problem is to arrange pairs of students in dorms so as to minimize social friction. As we can see, there are exponentially many combinations possible, not all students' constraints might be feasibly satisfied at once, and this is a hard problem. Now replace students with $N$ binary \emph{spins} each in $\sigma_i$ in $\{-1,1 \}$, arranged on a lattice in $\mathbb{R}^d$, and consider all their signed pairwise interactions $\sigma_i \sigma_j$. A \emph{configuration} is the data of states of $N$ spins $(\sigma_i)_{1 \leq i \leq N}$. If we wanted to associate a probabilistic model to each configuration, we could consider an energy model associated with a configuration $\sigma$'s energy
\begin{equation}\label{Gibbs}
    \mathbb{P}(\sigma) = \frac{\exp{\left[ -H(\sigma) \right] }}{Z}
\end{equation}
where $Z$ is a normalization constant called the \emph{partition function}. $\mathbb{P}$ is called the \emph{Gibbs measure} of the model. The object $H(\sigma)$, analogous to a log-likelihood, that represents the full energy of a configuration is central to our study as unique to the model considered, and called the \emph{Hamiltonian}. In the case just described, the Hamiltonian is :
\begin{equation}\label{Ising}
    H(\sigma) = - J \sum_{\langle i,j \rangle}{\sigma_i \sigma_j}
\end{equation}
where the $\langle i,j \rangle$ notation describes summation across nearest neighbors and $J$ is some constant absolute interaction strength. This is an example of the \emph{Ising model}. In general, the number of interacting students or particles is very large, and we are interested in the single minimum energy solution (ground state) observed at a macroscopic scale. Solving questions of interest around this model is done through the introduction of a parameter $\beta$ in (\ref{Gibbs}) which then becomes
\begin{equation}\label{GibbsT}
    \mathbb{P}(\sigma) = \frac{\exp{\left[ -\beta  H(\sigma) \right] }}{Z_{\beta}} , \qquad Z_\beta = \sum_{\sigma}{ \exp{\left[ -\beta  H(\sigma) \right] } }
\end{equation}
$\beta$ has an interpretation as an inverse temperature, and in the low-temperature limit, the log-sum-exp form of $Z$ turns into a hard maximum. The smooth maximum is studied through the \emph{free energy}
\begin{equation}
    F_{N}(\beta) = \frac{1}{N \beta} \mathbb{E}\Big[ \ln \sum_{\sigma}{\exp{\big(-\beta H(\sigma)\big) }} \Big]
\end{equation}
and its limit $F(\beta)$ as $N \rightarrow \infty$. Besides the Ising model, there are several models and Hamiltonians of interest, depending on their tractability and explanatory power for ferromagnetic properties.

\textbf{Spin glasses.} Besides summing pairwise interactions with constant coupling strength on binary-valued spins, there are several changes we might bring to the Hamiltonians we consider:
\begin{itemize}
    \item We can sum over \emph{all $(i,j)$ pairs}, turning the short-range interaction of (\ref{Ising}) into a mathematically more tractable \emph{mean-field model} (\cite{parisioverview}); 
    \item We can make the coupling strength $J$ random and i.i.d. for all $(i,j)$ pairs. For standard normal Gaussian $J_{i,j}$, this modification alongside the previous one gives rise to the Hamiltonian
    \begin{equation}
        H(\sigma) = -\frac{1}{\sqrt{N}} \sum_{i,j }{J_{ij} \cdot \sigma_i \sigma_j}
    \end{equation}
    which is the eminently studied \emph{Sherrington-Kirkpatrick} model (see \cite{panchenkointro} for an overview).
    \item We can consider interactions of order more than 2, introducing terms of the form $\sigma_{i_1} \cdots \sigma_{i_p}$.
    \item Finally, we can relax the magnetic interpretation of binary-valued spins and make them instead real-valued, but constrained to lie on a sphere. Combining these last two changes yields the Hamiltonian of the \emph{pure p-spin spherical spin-glass} 
    \begin{equation}
        H_p(\sigma) = -\frac{1}{N^{(p-1)/2}} \sum_{i,j }{J_{{i_1} \cdots {i_p}} \cdot \sigma_{i_1} \cdots \sigma_{i_p} }
    \end{equation}
\end{itemize}
where the spherical constraint usually is $\sum_{i}{\sigma_i^2}=N$. A \emph{mixed} spin glass, by opposition to a pure one, is then a linear combination of pure spin glasses of various degrees $p_i$.

Our interest in $p$-spin spherical spin-glasses comes from a result established by \cite{choromanska}: there exists a deep link between multi-layer perceptrons of depth $p$ and $p$-spin spherical spin glasses. More precisely, under very strong assumptions, the loss function of a ReLU multi-layer perceptron is the Hamiltonian of a spherical spin glass\footnote{Intuitively, introducing a measure on active information paths, where the ReLUs are 'dead' or not each layer, plays the role same as a Gibbs measure.}. The $\sigma_i$s play the role of the neural network weights, for a given realization of the initialization disorder $J$. This analogy is long-running and has proven very fruitful in giving both quantitative and qualitative results for the loss landscapes of neural networks and their critical points, for instance and using loose wording 'past a certain energy barrier, all local minima are equivalent' (\cite{essentiallynobarriers}) or 'the larger the network, the less the chance of finding poor minima'. The geometry of the loss landscape itself, and of the associated Gibbs measure, is a complex topic still under active investigation (\cite{littwin, subaggeometry, subaglandscapes,  subaglatest}).

\textbf{Weight decay and neural networks.} Spin glass physicists usually employ an added, external magnetic field that progressively tends to zero so as to make their calculations converge through \emph{symmetry breaking}, but the nature of that field is equivalent to $L^1$ regularization in neural network parlance. Weight decay or $L^2$ regularization is one of the more employed regularization techniques when training neural networks (\cite{smith}). Implementation details do affect its performance as measured by terminal accuracy, but combining it with adaptive gradient step size methods such as ADAM does give state of the art results (\cite{adamw}). However, despite the immediate mathematical formulation, reasons for its performance are not fully elucited (\cite{mechanisms}). In particular, $L^2$ regularization seems to interact with the layered structure of the loss landscape of neural networks themselves. Our work is similar in spirit to \cite{pratikloss}, but their modification is a multiplicative Gaussian noise added to networks' weights, the variance of which is annealed as training proceeds, according to a hyperparameter timescale $t_0$ that needs to be tuned. Our approach does not require timescale tuning, but rather, the initial strength of $L^2$ regularization. Similar ideas are also developed in \emph{sequentially tightening convex relaxations} (\cite{tightenafterrelax}) and \emph{homotopy continuation} (\cite{mobahi}). Recent work also shows that 'temperature is a proxy for weight constraints and regularization' (\cite{Martin}), so that simulated annealing can actually be achieved by regularization annealing.

\section{Complexity of spin glasses}
\subsection{From Kac-Rice to Bray and Dean}

{\bf Exponential complexity.} Denoting by $\mathcal{N}(\alpha, \epsilon)$ the number of critical points of energy $\epsilon$ and (normalized) Hessian index $\alpha$, we look at the \emph{complexity} or normalized logarithmic number $\Sigma(\alpha, \epsilon)$ defined as follows
\begin{equation*}
\langle \mathcal{N}(\alpha, \epsilon) \rangle = \exp \left[ N \cdot \Sigma(\alpha, \epsilon )\right]
\end{equation*}

or equivalently (this formulation also looks like a \emph{good rate function} in large deviations theory)
\begin{equation}
\Sigma(\alpha, \epsilon ) = \frac{\ln \left( \langle \mathcal{N}(\alpha, \epsilon) \rangle \right)}{N}
\end{equation}
As observed by \cite{dauphin}, for the purposes of studying which critical points are minima as opposed to simply saddle points, independent-sign eigenvalues would have an exponentially vanishing (in $N$) probability of all having the same sign. This justifies being interested in the \emph{exponential} decay rate.

{\bf Generalized Kac-Rice.} The Kac-Rice formula enables us to compute explicitly the number of crossings or critical points of a given function. Generalized to an N-dimensional scalar random field $\phi$ (\cite{fyodorov}), it reads that the number of \emph{minima} of $\phi$ of energy $\epsilon$ and index $\alpha$ is exactly
\begin{equation}\label{kacrice}
\mathcal{N}(\alpha, \epsilon) = \int_{V}{dx \prod_{i=1}^{N}{ \delta(\partial_{i} \phi(x) ) |\det{H(x)}| \delta(I(H(x))-N \alpha) \delta(\phi(x) - N \epsilon)   }}
\end{equation}
$H(x)$ is the Hessian matrix of field $f$ at point x, made of elements $H_{i,j}(x) = \partial_{i} \partial_{j} \phi(x)$. $I(H)$ is the index of $H$, defined as its number of negative eigenvalues.

Crucially, we also assume that $\phi$ is a Gaussian field, isotropic (rotation-invariant), with zero average, so that it is given completely by its covariance function $f$ :
\begin{equation*}
\langle \phi(x) \phi(y) \rangle = N f(\frac{(x-y)^2}{2N})
\end{equation*}

These assumptions are effectively verified in the case of Hamiltonians of large-scale, spherical spin glasses with Gaussian disorder (\cite{brayanddean}). In fact, isotropic Gaussian functions on the $N$-dimensional sphere are \emph{exactly} mixed p-spin spherical spin-glasses \cite{auffingercomplexity}; this is seen by expanding covariance function $f$ in power series, where monomial terms are exactly pure spin glasses. After long calculations that involve relaxing the Dirac distributions in (\ref{kacrice}) as infinitesimal variance Gaussians, so that $\delta(x) = \lim_{\epsilon \rightarrow 0}{e^{-x^2/(2\epsilon) }}$, and looking to compute $\mathbb{E}\left[ \mathcal{N}(\alpha, \epsilon) \right]$ (we note following \cite{subagcomplexity} that it is enough to study the first moment of these quantities as they converge to their average a.s.), we are left with computing an integral involving the absolute value of a determinant over random matrices in the GOE (\cite{fyodorov}), which ultimately leads to closed-form or tractable formulas.

In what follows, we take this triple perspective of Gaussian initialized feedforward neural networks, spin glasses, and random fields, assume it is valid, and examine the theoretical implications of this analogy.

\subsection{Relations between loss and Hessian spectrum}

Because of the Gaussian character of our analysis, the Hessian of random field $\phi$ is itself a Gaussian random matrix, so that its eigenvalues asymptotically follow the Wigner semi-circle law. Hence there is an implicit relation linking $\alpha$ (the fraction of negative eigenvalues) and $\xoverline{\lambda}$ (the average eigenvalue $\xoverline{\lambda} = \int{ \lambda \rho(\lambda) d\lambda}$ for the eigenvalue density $\rho$) :

\begin{eqnarray}\label{arccos}
\alpha = \frac{2}{\pi} \int_{ \frac{\xoverline{\lambda}}{2\sqrt{f''(0)} } }^{1}{\sqrt{1-t^2} dt} = \frac{1}{\pi} \left[ \  \arccos \frac{\xoverline{\lambda}}{2\sqrt{f''(0)}} - \frac{\xoverline{\lambda}}{2\sqrt{f''(0)}} \sqrt{1- \frac{\xoverline{\lambda}^2}{4 f''(0)} } \right]
\end{eqnarray}
also seen as an index constraint - for Gaussian random functions, controlling the worst-case positive eigenvalue is the same as controlling the average eigenvalue.
We introduce constants $P$ and $Q$ as limits when $N \rightarrow \infty$ of
\begin{eqnarray*}
P_N = (1- \frac{2}{N}) \frac{f(0)}{f''(0)} + \frac{f'(0)^2}{f''(0)^2} \\
Q_N = (1+ \frac{2}{N}) \frac{f(0)}{f''(0)} - \frac{f'(0)^2}{f''(0)^2}
\end{eqnarray*}
Now, in the specific case of random Gaussian fields/spin glasses, we have the most likely average value conditional on $\epsilon$:
\begin{equation}\label{mostlikelyepsilon}
\xoverline{\lambda}(\epsilon) = \frac{2 f'(0)}{f''(0) P} \epsilon
\end{equation}

so that introducing the \emph{critical energy} $\epsilon_c$ as $\epsilon_c = P \frac{f''(0)^{3/2}}{f'(0)}$, we can rearrange the previous as
\begin{equation}\label{invertedwigner}
\alpha(\epsilon) = \frac{1}{\pi} \arccos \frac{\epsilon}{\epsilon_c} - \frac{\epsilon}{\pi \epsilon_c} \sqrt{1- (\frac{\epsilon}{\epsilon_c})^2} = F(-\frac{\epsilon}{|\epsilon_c|})
\end{equation}
with $F$ the inverse integrated Wigner distribution, Taylor expanding as
\begin{eqnarray}\
\alpha(\epsilon) &\underset{\epsilon \rightarrow 0}{\sim} \frac{1}{2} - \frac{2}{\pi} \cdot \frac{\epsilon}{\epsilon_c} \\
\alpha(\epsilon) &\underset{\epsilon \rightarrow \epsilon_c}{=} O\Big( (\epsilon - \epsilon_c )^{3/2} \Big)
\end{eqnarray}
that recovers the $3/2$ power scaling seen in other works (\cite{pennington}). 

\begin{figure}
    \includegraphics[width=0.33\textwidth]{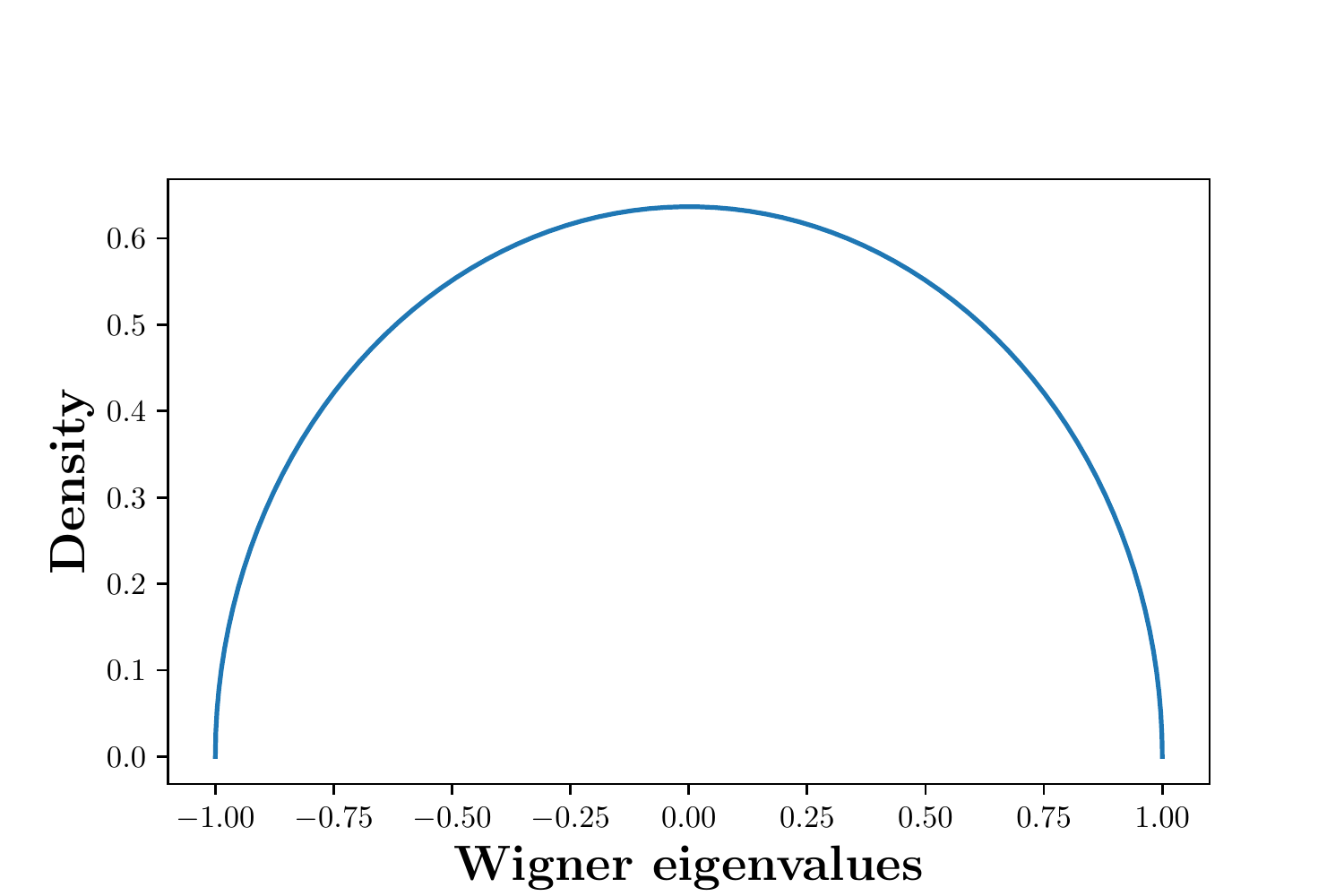}
    \includegraphics[width=0.33\textwidth]{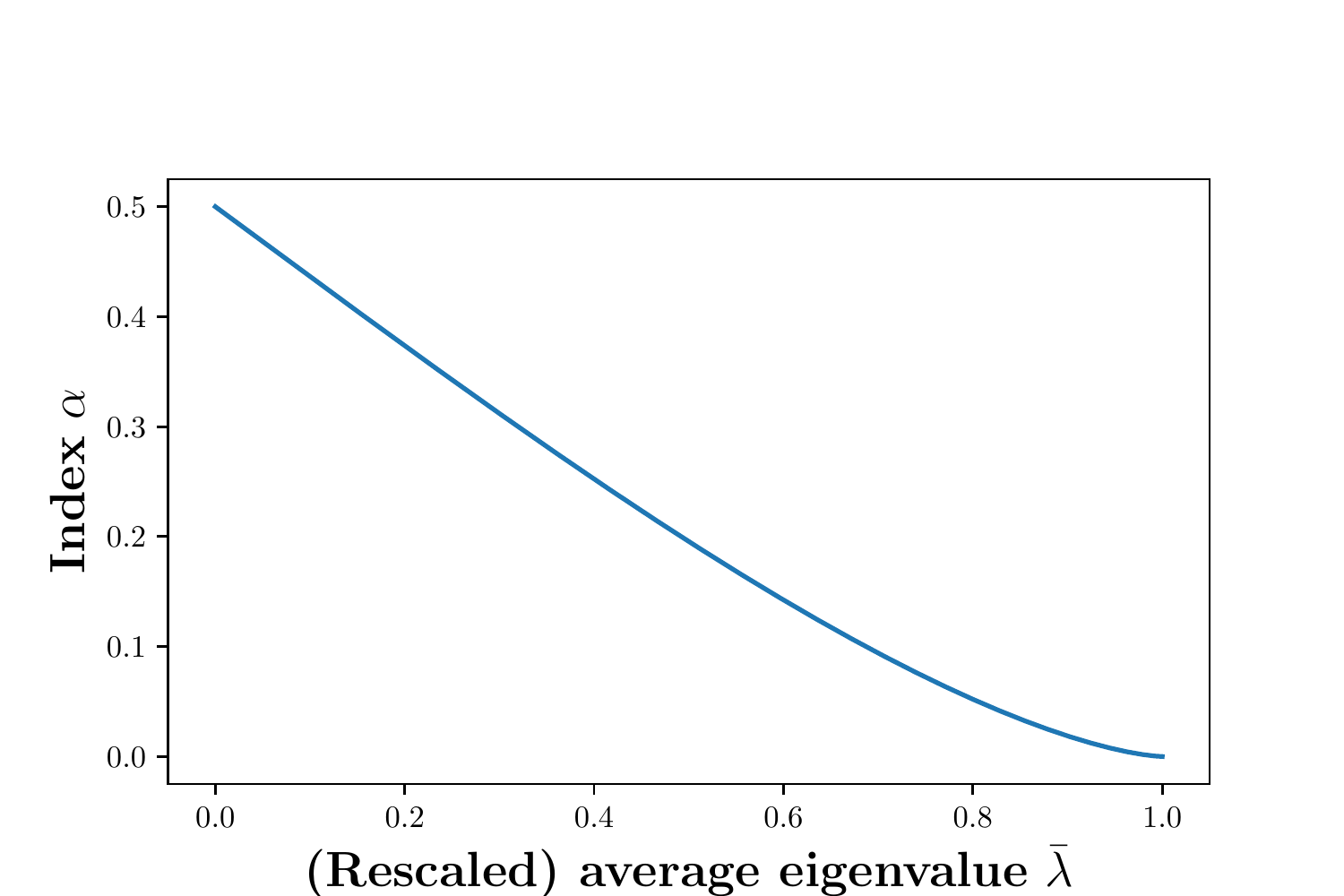}
    \includegraphics[width=0.33\textwidth]{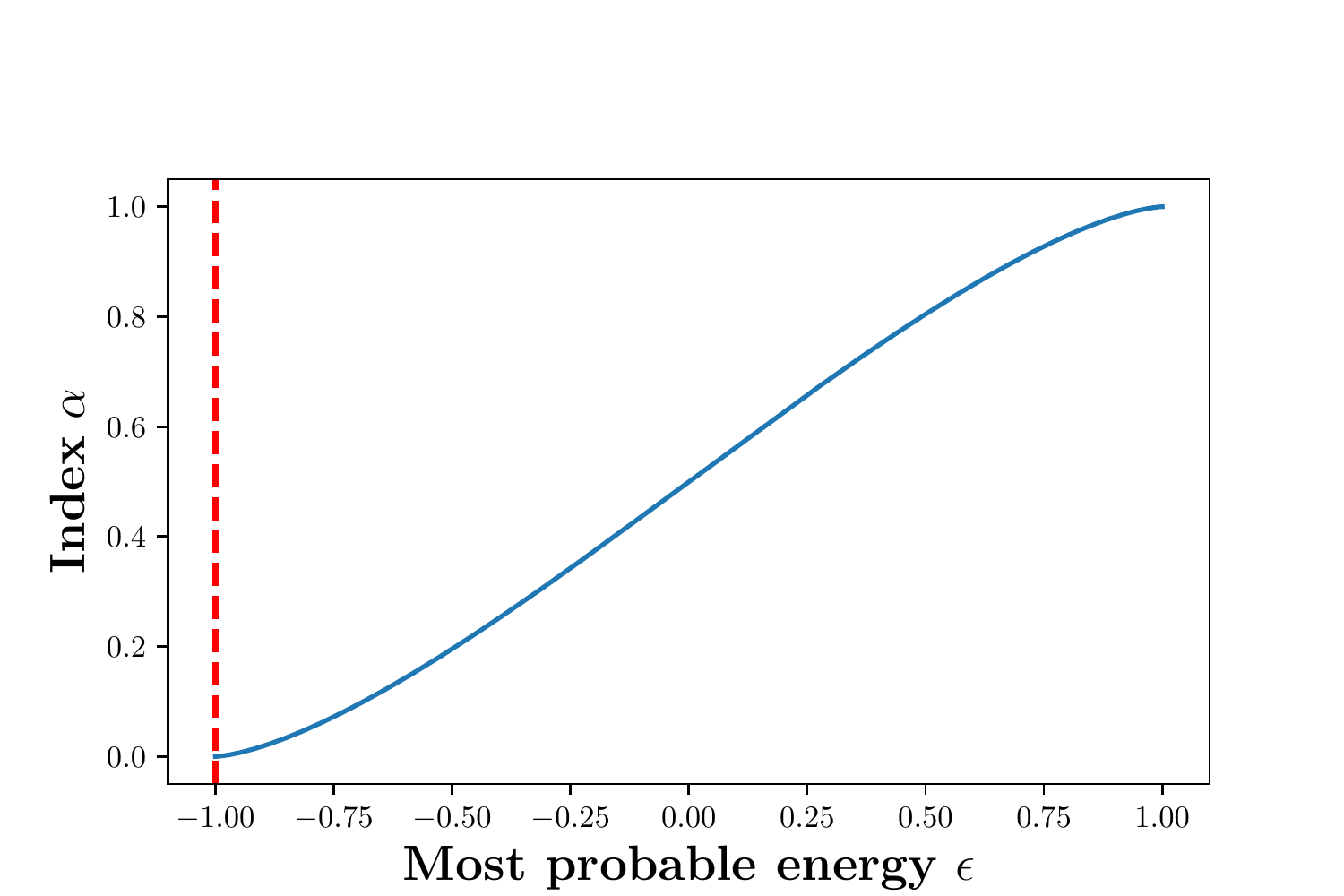}
    \caption{\textbf{Eigenvalues and index distributions of interest.} \emph{Left}: the classical, Wigner semicircle law asymptotically followed by the Hessian eigenvalues. \emph{Middle}: Inverting this integrated relation (see (\ref{arccos})) yields the most likely index $\alpha$ as a function of $\xoverline{\lambda}$. \emph{Right}: Same relation holds reverted (\ref{invertedwigner}) with respect to most likely energy $\epsilon$. Dashed red line is $\epsilon_c<0$.}
    \label{fig:inverseindex}
\end{figure}

\newpage
{\bf Quadratic log-complexity.} Finally we get, to leading order, the complexity:

\begin{theorem}[Thm~4 in~\cite{brayanddean}]
\begin{align*}
\Sigma(\alpha, \epsilon) &= -\frac{1}{2 f''(0) Q} [ \epsilon^2 -2 \frac{f'(0)}{f''(0)} \epsilon \xoverline{\lambda} + \frac{P}{2}\xoverline{\lambda}^2 ] + \frac{1}{2} \ln \left( \frac{f''(0) V^{2/N} } {2 \pi e |f'(0)|} \right) \\
&=  -\frac{1}{2 f''(0) Q} \cdot \psi(\alpha, \epsilon) + \cdots
\end{align*}
\end{theorem}
itself an affine function of the only relevant part, that we name $\psi$
\begin{eqnarray}\label{psi}
\psi(\xoverline{\lambda}(\alpha), \epsilon) = \epsilon^2 -2 \frac{f'(0)}{f''(0)} \epsilon \cdot \xoverline{\lambda}(\alpha) + \frac{P}{2}\xoverline{\lambda}(\alpha)^2
\end{eqnarray}
so that, introducing $\bm{u} = (\epsilon, \xoverline{\lambda}(\alpha))$:
\begin{eqnarray}
    \Sigma(\xoverline{\lambda}, \epsilon) = \Sigma(\bm{u}) = C_1 - \frac{1}{2} C_2 \bm{u}^T \bm{M} \bm{u} \\
    \bm{M} = \begin{pmatrix}
1 & -\frac{f'(0)}{f(0)} \\
-\frac{f'(0)}{f(0)} & \frac{1}{2}P\\
\end{pmatrix}
\end{eqnarray}
This is a diagonalizable quadratic form in $(\epsilon, \xoverline{\lambda})$. In neural network terms with the weights of our network denoted as $\theta$, $\epsilon$ is the level of our loss function $\epsilon_{\theta}$, and $\xoverline{\lambda}_\theta$ our average Hessian eigenvalue (or directly, regularization strength for centered networks) at given index $\alpha_\theta$. $\bm{M}$ is a constant matrix depending only on the network architecture, through derivatives of its input covariance function $f$ at 0.

\subsection{Complexity gradient descent}

\textbf{Interpretation.} From a complexity perspective, it is strictly equivalent to move $\epsilon$ (by taking a stochastic gradient descent step) or $\xoverline{\lambda}$ (by just shifting the $L^2$-regularization parameter) as they play symmetric roles. We therefore propose a double interpretation of regularized minibatch gradient descent:
\begin{itemize}
\item gradient descent steps, with a certain probability, identify further descent directions that fold the negative part of the Hessian eigenspectrum back into the positive ($\alpha$ increases). The shape of that spectrum is rigid (Wigner) and shifts.
\item This folding lowers the error level $\epsilon$, but also simultaneously shifts the average eigenvalue $\xoverline{\lambda}(\epsilon)$ right, according to equation (\ref{mostlikelyepsilon}) as $f'(0)<0$. This is equivalent to saying $Tr \bf{H}$, or $\Delta \bf{H}$ (the \emph{Laplacian} of $\bf{H}$) shifts right.
\item Shifting the average eigenvalue right can also be achieved by $L^2$ regularizing the loss function. This comes at the expense of minima precision. Generally this is not an issue, as the strength of $L^2$ regularization is small and constant.
\end{itemize}

\textbf{Link with $L^2$ regularization.} $L^2$ regularization with a parameter $\lambda_0$ enforces the constraint $\xoverline{\lambda} \geq \lambda_0$. At one extreme of the $(\epsilon, \xoverline{\lambda})$ spectrum, where there is no regularization, we would converge exactly to the right minimum if we knew which one to pick from a multitude of those. Conversely, at the other end, with extreme regularization, we can hope to converge faster to a single minimum, but it is now far away from the 'true' minimum. This quantifies the precision/speed tradeoff in SGD optimization, and suggests to match the regularization parameter schedule during the training of neural networks to the learning rate decay schedule, since (\ref{mostlikelyepsilon}) also gives:
\begin{equation}\label{percentages}
    |\frac{d \epsilon}{\epsilon}| = |\frac{d \xoverline{\lambda}}{\xoverline{\lambda}}|
\end{equation}
sidestepping the difficulty of estimating constants like $P$, $f'(0)/f''(0)$ and $Q$, which are hardly accessible. For instance, $\frac{d \epsilon}{\epsilon}$ can be estimated, or derived from traditional decay schemes for learning rate, such as the typically used power law :
$$\epsilon_t \sim \epsilon_0 \cdot \frac{1}{\Big( 1+ t \Big)^\eta } $$ with generally $\eta = \frac{1}{2}$.

\textbf{Complexity as a surrogate objective.} Complexity includes information both from the level of the loss and the curvature of the landscape - the latter a desirable feature as an indication of the flatness of minima and therefore a trained network's generalization ability. Following the gradient of the complexity makes sense if only to try and locate the unique \emph{ground state} of our optimization problem, assuming it exists. But another argument is the fact that most critical points are to some extent saddle points and that traditional gradient descent algorithms will slow down their trajectory in their vicinity (\cite{dauphin}) according to Kramers law (\cite{pratikloss}) - therefore making the path with the least intermediate critical points potentially the fastest.

The gradient of complexity $\Sigma(\bf{u})$ is known in closed form, up to constants, as 
\begin{equation}\label{gradient}
\bf{\nabla} \Sigma(\bf{u}) = - \bf{M} \bf{u}
\end{equation}
In practice, both the estimation of the $f$-dependent constants in $\bf{M}$, and the $\xoverline{\lambda}$ gradient steps are hurdles that we address in the Algorithms section. 

\subsection{Match with stylized facts}
The picture that our complexity gradient paints is compelling, as it matches numerous empirical \emph{stylized facts} broadly consistent with separate observations in the literature:
\begin{itemize}
    \item a picture compatible with the remarks in \cite{mechanisms}, namely that when using batch normalization (which enforces spherical conditions on layers' input data), $L^2$ regularization works by increasing the effective learning rate of stochastic gradient descent steps - a direct consequence of eq. (\ref{psi}) as.
\begin{equation}
    \frac{\partial \psi}{\partial \epsilon}|_{\epsilon} = \epsilon \Big( 1 + \frac{2}{1+ \frac{f(0)f''(0)}{f'(0)^2}} \Big) > \epsilon
\end{equation}
    \item need for an $L^2$ regularization coefficient the same order of magnitude as the learning rate (\cite{smith});
    \item exponential number of critical points consistent with robustness of gradient descent methods to initialization;
    \item increasing curvature in the loss landscape as one mechanism easing training (\cite{dauphin, Martin});
    \item potential clustering in critical points, amenable to warm restart methods (\cite{sgdr});
    \item multiple local minima (\cite{essentiallynobarriers});
    \item existence of power laws and phase transitions (\cite{deeplearningscaling}) during training across multiple domains.
\end{itemize}
Those have been observed independently of the Gaussian character of initialization, which suggests some universal mechanisms beyond the scope of this paper might be at play.
\section{Algorithms}
In this section we describe three algorithms that implement the idea of complexity gradient descent. Those perform approximations as required in practice.  Constants like $P$, $f'(0)/f''(0)$ and $Q$ that appear in matrix $\bf{M}$ in (\ref{gradient}) are hardly accessible. 

\textbf{Baseline algorithm.} Based on equation (\ref{percentages}), one way to enforce rough proportionality without having to numerically estimate constants is to make the regularization strength decay strictly proportional to the loss decay :
    $\frac{\lambda_i}{\lambda_0} = \frac{L(i)}{L(0)}$
with $L$ the training loss of the network for epoch $i$. The baseline algorithm is proposed as Algorithm 1.

\begin{algorithm}[H]
	\caption{Complexity Gradient Descent - Matched Annealing.}
	\begin{algorithmic}
	\State Assume initial $L^2$ regularization strength $\lambda_0$
    \State Initialize network weights $\theta$ randomly (Gaussian i.i.d.) on the $N$-sphere
    \State Record $L(0)$ and take $K$ further, initial gradient descent steps, recording $L(i), \quad i \leq K$
    \State For each epoch $i$:
    \Repeat
    \State Reset gradients: $d\theta \gets 0$
    \State Set $\lambda_i = \lambda_0 \cdot S_K(L(i) / L(0))$, where $S_K$ is a smoothing operator with memory $K$
    \State Set $f_{i}(\theta) = L(\theta) + \lambda_i \cdot || \theta || ^2 $
    \State Perform a $\theta$ gradient step w.r.t. $f_{i}(\theta)$
    \State Project back weights $\theta$ onto the $N$-sphere
    \State $i \leftarrow i+1$
    \Until $N_{epochs}-i$
	\end{algorithmic}
    
\end{algorithm}
Different smoothing schemes $S_K$ are possible (the trivial one being the identity function). In practice, Algorithm 1 is not robust to the addition of a constant to $L$, which might make it unsuitable from a theoretical perspective; this can be prevented for instance by using exponential smoothing. Furthermore, different smoothing schedules $S$ can be considered and provide robustness-adapatability tradeoffs. In order to remedy this and leverage the intuition we've gotten from physical insights, we propose two further algorithms.

\textbf{Cosine annealing.}  \cite{sgdr} propose to anneal the learning rate in SGD following a restarted cosine schedule, in order to accelerate convergence. Similarly, this can be mirrored in the $L^2$ regularization strength schedule so as to do \emph{warm restarts in curvature}, as per Algorithm 2. 
\begin{algorithm}[H]
	\caption{Complexity Gradient Descent - Cosine Annealing.}
	\begin{algorithmic}
	\State Assume initial $L^2$ regularization strength $\lambda_0$
	\State Assume regularization strength restarting period $T$
    \State Initialize network weights $\theta$ randomly (Gaussian i.i.d.) on the $N$-sphere
    \State Record $L(0)$
    \State For each epoch $i$:
    \Repeat
    \State Reset gradients: $d\theta \gets 0$
    \State Set $\lambda_i = \lambda_0 \cdot (1+ \cos(2 \pi \cdot i / T))$
    \State Set $f_{i}(\theta) = L(\theta) + \lambda_i \cdot || \theta || ^2 $
    \State Perform a $\theta$ gradient step w.r.t. $f_{i}(\theta)$
    \State Project back weights $\theta$ onto the $N$-sphere
    \State $i \leftarrow i+1$
    \Until $N_{epochs}-i$
	\end{algorithmic}
    
\end{algorithm}

\textbf{Two-step annealing.} Using the insight that for $\lambda_1 \approx \lambda_2$, gradient descent weights for $L(\theta) + \lambda_1 \cdot || \theta || ^2$ are an excellent starting point to gradient descent on $L(\theta) + \lambda_2 \cdot || \theta || ^2$, we propose to implement an EM-like algorithm for complexity with two standard $\theta$ gradient steps of different sizes. This gives rise to Algorithm 3:
\begin{algorithm}[H]
	\caption{Complexity Gradient Descent - Two-step Annealing.}
	\begin{algorithmic}
	\State Assume initial $L^2$ regularization strength $\lambda_0$ and schedule $\lambda_t$
    \State Initialize network weights $\theta$ randomly (Gaussian i.i.d.) on the $N$-sphere
    \State Record $L(0)$
    \State For each epoch $i$:
    \Repeat
    \State Reset gradients: $d\theta \gets 0$
    \State Set $f(\theta) = L(\theta) + \lambda_i \cdot || \theta || ^2 $
    \State Perform a $\theta$ gradient step of size $\eta$ w.r.t. $f(\theta)$ \text{(complexity E step)}
    \State Perform a $\theta$ gradient step of size $(\lambda_i - \lambda_{i-1}) \cdot \eta$ w.r.t. $f(\theta)$ \text{(complexity M step)}
    \State Project back weights $\theta$ onto the $N$-sphere
    \State $i \leftarrow i+1$
    \Until $N_{epochs}-i$
	\end{algorithmic}
    
\end{algorithm}

Finally, we note that under the Gaussian-Wigner Hessian assumption that it would theoretically be possible to derive numerically the smallest $L^2$ regularization strength that, in expectation, convexifies the loss (that would be $\alpha(\epsilon)$ under our notations) ; while this opens up the possibility of using convex optimization methods, we leave this open for further research. Importantly, we also leave practical evaluation of these algorithms at scale as future work, which will come in Part II.

\section{Discussion and further work}
Using results from \cite{brayanddean} and \cite{auffingercomplexity}, we have introduced \emph{complexity gradient descent}, which posits that the complexity of critical points of the losses of neural networks can act as a surrogate objective for optimization that takes into account both the level of the loss and also the quality (flatness) of minima. This paper has focused on the reinterpretation of statistical physics arguments, and shown that they match most of the stylized facts observed during the process of training deep neural networks. We have also proposed heuristic algorithms taking advantage of the newly uncovered symmetry between learning rate and strength of $L^2$ regularization. A second, companion paper will examine the empirical performance of these algorithms at scale. Further work will also hopefully look to extend these results away from the purely Gaussian case, using recent \emph{universality} results from spin glass mathematics and random matrix theory. Finally, and maybe most intriguingly, all of these results involve constants $f(0), f'(0), f''(0)$ that strictly depend on the Taylor expansion in $0$ of the covariance of $\phi$, that is, constants that could possibly be estimated purely by forward passes before a network is trained. This is consistent with some results anecdotally reported in the deep learning literature and represents one of the most intriguing ideas coming out of this line of work, with potential applications to further optimization methods and neural architecture search.

\section{Acknowledgements}
The authors would like to thank Bilal Piot and Mohammad Gheshlaghi Azar for helpful comments and suggestions.

\nocite{panchenkointro,pratikloss,subaglatest,subaglandscapes,subaggeometry,subagcomplexity,fyodorov}
\nocite{auffingerrmt,auffingercomplexity,parisioverview,brayanddean,dauphin,choromanska,pennington,spectraluniversality,gabrie, mechanisms, adamw, mobahi, pmlr-v28-sutskever13, Adam, smith, duSGD, zouSGD, littwin}

\bibliographystyle{iclr2018_conference}
\bibliography{iclr2018_conference}

\end{document}